# A Method for Using Belief Networks as Influence Diagrams

Gregory F. Cooper
Medical Computer Science Group
Medical School Office Building
Stanford University
Stanford, CA 94305

**ABSTRACT**

This paper demonstrates a method for using belief-network algorithms to solve influence-diagram problems. In particular, both exact and approximation belief-network algorithms may be applied to solve influence-diagram problems. More generally, knowing the relationship between belief-network and influence-diagram problems may be useful in the design and development of more efficient influence diagram algorithms.

## 1. Introduction

A belief network is an acyclic, directed graph that represents the probabilistic dependencies among a set of chance variables (see Pearl86a for a detailed discussion). Belief-network algorithms perform probabilistic inference on belief networks. Figure 1 shows a simple, abstract belief network. The absence of an arc from A to C indicates that node C is conditionally independent of node A given the value of node B. The calculation of $P(A=T \mid C=T)$ is an example of a belief network inference problem; the solution in this case is 0.62.

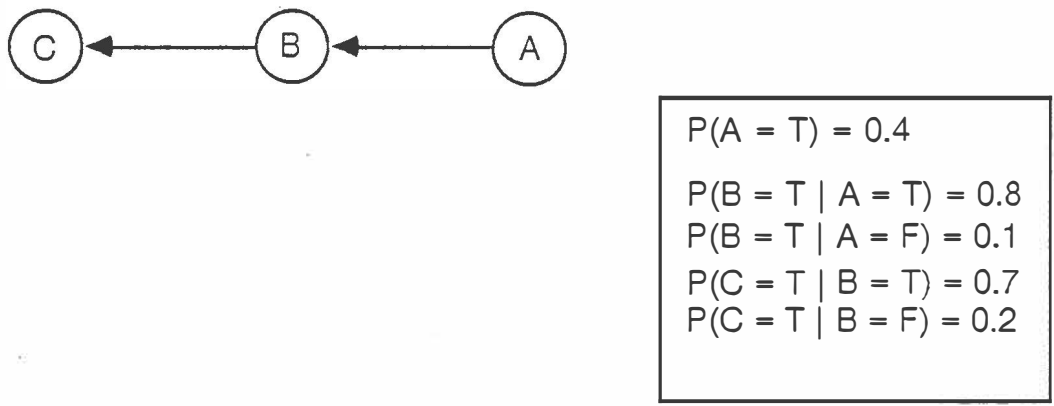

$P(A = T) = 0.4$

$P(B = T \mid A = T) = 0.8$
$P(B = T \mid A = F) = 0.1$
$P(C = T \mid B = T) = 0.7$
$P(C = T \mid B = F) = 0.2$

Figure 1. A example of a simple belief network. A circle represents a chance node.

An influence diagram is a belief network that is augmented with decision nodes and a value node (see Howard84 and Shachter86a for a detailed discussion). The relationships among chance nodes, decision nodes, and a value node are explicitly represented in the network of the influence diagram. A primary task of an influence-diagram inference system is the determination of the decision alternatives that maximize expected value. We focus on this task in this paper, although extensions to other inference tasks, such as calculating value-of-information and sensitivity analysis, are possible using slight modifications of the techniques we present below.

Figure 2 is an example of an influence diagram that has been generated by the addition of binary decision-node $D_1$ and a binary value-node V to the belief network in Figure 1. The arc from C to $D_1$ indicates that the value of C will be available when decision $D_1$ is made. In other words, A and B are hidden variables when decision $D_1$ is made, whereas the value of C will be available as evidence. The arcs from A and $D_1$ to V indicate that the value of the state of the diagram is a function of A and $D_1$. Determining the value of node $D_1$ (i.e., $Action_1$ or $Action_2$) that maximizes the expected value of the model in light of available evidence C is the primary influence-diagram decision problem we consider here. Suppose that C=T; then, it so happens that $Action_1$ is the solution to the decision problem in Figure 2, and the expected value of taking $Action_1$ is 2.1.

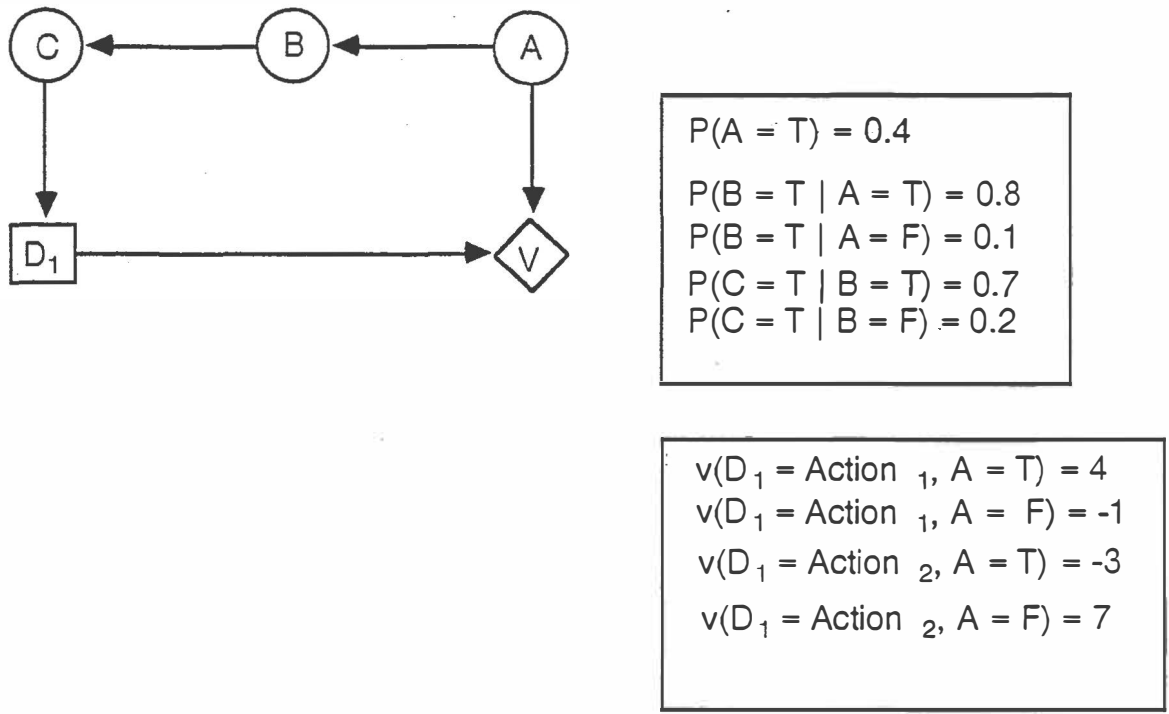

Figure 2. An example of a simple influence diagram. A circle represents a chance node, a square represents a decision node, and a diamond represents a value node.

A well-known influence diagram algorithm has been developed by Shachter [Shachter86a, Shachter87b] which also can be used as a belief network algorithm [Shachter86b]. Conversely, Shachter has shown that a belief network algorithm (i.e., a probabilistic influence-diagram algorithm, in his terminology) of a particular type (i.e., that uses arc reversal and node removal operations) can be used to solve influence-diagram problems [Shachter86b, Shachter87a]. The remainder of this paper builds on Shachter's results by providing a mapping from influence diagrams to belief networks that differs from his mapping in several key respects. First, the transformation in this paper is applicable to any belief network algorithm, rather than assuming that a belief-network algorithm uses arc-reversal and node-elimination operations. In particular, the general transform covered here allows the application of any exact, approximation, or heuristic belief-network algorithm in solving influence-diagram problems. Second, the transformation in this paper uses recursion to solve a decision problem without requiring the storage of potentially large tables of intermediate results. Third, our formulation allows the instantiation of chance-variables before a decision problem is solved; this capability is important in the use of influence diagrams in expert systems, where background information (i.e., instantiated chance-variables) often is available when a system is asked to make a decision recommendation.

## 2. Transforming an Influence Diagram to a Belief Network

The basic steps in transforming a generic influence diagram ID that is oriented and regular (for formal definitions see [Shachter86a], page 875) to a belief network BN involve converting decision nodes to chance nodes, converting the value node to a chance node, and transforming the value function to a probability function. Let D be the set of all n decision nodes in ID. A regular influence diagram specifies a total order on all the decision nodes in D. This total order is used to determine the time precedence in which decisions must be made. For notational convenience, assume that the decision nodes are relabeled if necessary such that the total order is $D_1, ..., D_n$. Let L be a list with the structure $((D_1, \Pi_{D_1}, \Pi'_{D_1}), (D_2, \Pi_{D_2}, \Pi'_{D_2}), ..., (D_n, \Pi_{D_n}, \Pi'_{D_n}))$, where $\Pi_{D_i}$ is a list of all the nodes in ID with information arcs into $D_i$ (including implicit no-forgetting arcs), and $\Pi'_{D_i}$ is the set of all chance nodes in $\Pi_{D_i}$. After list L is constructed, the arcs into each $D_i$ are deleted from ID. Each decision node $D_i$ in ID is converted to a chance node, and is relabeled if necessary to maintain consistency with the labeling of chance nodes already in ID. Each decision alternative of each chance node $D_i$ is assigned some prior probability that is greater than 0 and less than 1 such that the prior probabilities of the alternatives for a given $D_i$ sum to 1; as discussed in Section 3, each $D_i$ will always be instantiated, and therefore the specific prior-probability assignment to $D_i$ is not critical.

Without loss of generality, we will consider only the case in which an influence diagram has a single value node V. Let $v(\Pi_v)$ denote the value function in ID, where $\Pi_v$ represents those variables (nodes) in ID with arcs to value node V. We will transform function v to a probability function. Define node V to be binary-valued with possible values T and F. The probability function relating nodes in $\Pi_v$ to node V is defined as follows:

$$P(V = T \mid \Pi_v) = \frac{[v(\Pi_v) + k_2]}{k_1} \tag{1}$$

$$\text{where } k_1 = (\max_{\Pi_v}[v(\Pi_v)] - \min_{\Pi_v}[v(\Pi_v)])$$

$$\text{and } k_2 = - \min_{\Pi_v}[v(\Pi_v)]$$

In essence, Equation 1 linearly maps the range of v from an arbitrary subset of an interval of the real numbers to a subset of the real numbers that is inclusively between 0 and 1. Figure 3 shows the belief network representation of the influence diagram in Figure 2.

57

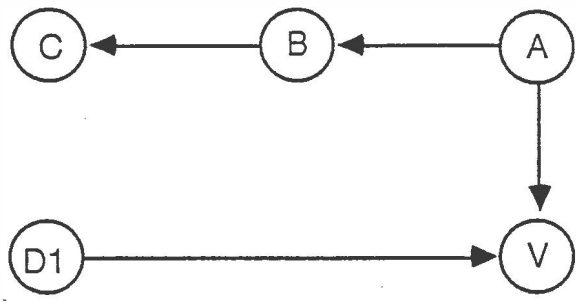

$P(A = T) = 0.4$

$P(B = T \mid A = T) = 0.8$
$P(B = T \mid A = F) = 0.1$
$P(C = T \mid B = T) = 0.7$
$P(C = T \mid B = F) = 0.2$
$P(V = T \mid D_1 = Action_1, A = T) = 0.7$
$P(V = T \mid D_1 = Action_1, A = F) = 0.2$
$P(V = T \mid D_1 = Action_2, A = T) = 0.0$
$P(V = T \mid D_1 = Action_2, A = F) = 1.0$

$k_1 = 10$
$k_2 = 3$
$L = ((D_1 \ (C) \ (C)))$

Figure 3.  A belief-network representation of the influence diagram in Figure 2.

## 3. Solving an Influence-Diagram Problem Using Its Belief-Network Representation

3.1 Solving a Single-Decision Problem

Let E be a set containing chance variables in BN with known values (i.e., evidence), including those decisions $D_1, ..., D_{n-1}$ in D that have already been made. Let $D_n$ represent the remaining decision to be made. At this stage in the decision-making process assume that all variables in $\Pi_{D_n}$ are instantiated and are therefore in E. The influence diagram inference problem at this point is to determine the instantiation of $D_n$ that produces a maximization of expected value (MEV):

$$MEV(D_n, E) = \max_{D_n} [ \sum_{\Pi'_v} v(\Pi_v) \, P(\Pi'_v \mid D_n, E) ] \qquad (2)$$

Equation 2 determines the instantiation of $D_n$, termed $D^*_n$, that maximizes expected value in light of evidence E. We can replace $v(\Pi_v)$ in Equation 2 by an equivalent function derived from Equation 1:

$$\begin{aligned}
MEV(D_n, E) &= \max_{D_n}[\sum_{\Pi'_v} v(\Pi_v) P(\Pi'_v | D_n, E)] \\
&= \max_{D_n}[\sum_{\Pi'_v} (k_1 P(V = T | \Pi_v) - k_2) P(\Pi'_v | D_n, E)] \\
&= k_1 \max_{D_n}[P(V = T | D_n, E)] - k_2
\end{aligned} \quad (3)$$

Equation 3 illustrates, that to maximize expected value of a single decision, it is sufficient to determine the instantiation of the decision variable $D_n$ that maximizes the probability $P(V=T | D_n, E)$. The calculation of $P(V=T | D_n, E)$ for a given instantiation of $D_n$ is the standard calculation performed by belief-network inference algorithms. Therefore, the overall inference mechanism involves applying a belief-network algorithm to calculate $P(V=T | D_n, E)$ for each instantiation of $D_n$ and then choosing the instantiation that produces the maximal value of $P(V=T | D_n, E)$. This instantiation is equivalent to $D^*_n$ calculated by applying an influence-diagram algorithm to the corresponding influence-diagram problem. For example, the solution to the decision problem in Figure 3 is $D^*_1$ = Action$_1$, which as expected is the same decision alternative calculated previously using the influence diagram in Figure 2. In particular, $P(V = T | D^*_1, C = T) = 0.51$, and by Equation 3 the expected value of taking Action$_1$ is $k_1 \times 0.51 - k_2 = 10 \times 0.51 - 3 = 2.1$, as before.

### 3.2 Solving a Multiple-Decision Problem

Let E be a set containing chance variables in BN with known values (i.e., evidence), including those decisions $D_1, ..., D_{i-1}$ in D that have already been made. Let $D_r = D_i, ..., D_n$ be an ordered list of the remaining decisions to be made. Among the primary influence-diagram inference problems is the determination of the instantiation of $D_i$ that produces a maximum expected value. To determine the maximum expected value, the following steps are taken. All uninstantiated chance nodes are removed from $\Pi_{D_i}$ and $\Pi'_{D_i}$, because decision $D_i$ must be made only in light of currently available information. Alternatively, uninstantiated nodes in $\Pi_{D_i}$ and $\Pi'_{D_i}$ may be instantiated to desired values in order to determine the optimal decision $D_i$, contingent on those instantiated values existing. Equation 4 is a recursive version of Equation 3 that determines the maximum expected value based on the optimal decision value for decision $D_i$ in light of evidence E, when called with $MEV(D_r, E)$. The arg-max version of the function in Equation 4 can be used similarly to determine the optimal decision value of decision $D_i$ in light of evidence E. The optimal decision values for each of the future decisions $D_{i+1}, ..., D_n$ can be computed, contingent on the values of each of their information predecessors, by storing intermediate calculations of the recursive function in Equation 4.

$$MEV(d, e) = k_1 f(d, e) - k_2$$

where (4)

$$f(d, e) = \max_{h[d]} [ \sum_{\Pi'_{h[t[d]]}} f(t[d], e \cup \Pi_{h[t[d]]}) P(\Pi'_{h[t[d]]} | h[d], e) ]$$

and where h[d] returns the first element of d (i.e., in LISP notation, h[d] = (CAR d)), t[d] returns the list that remains when the first element of d is removed (i.e., in LISP notation, t[d] = (CDR d)), $\emptyset$ = the empty list, $\Pi'_\emptyset = \Pi_V$, $f(\emptyset, e) = P(V = T | \Pi_V)$, and $P(\emptyset | h[d], e) = 1$.

Equation 4 provides a constructive inductive proof that MEV(d, e) computes the maximum expected value based on the optimal decision value for decision $D_i$. In essence, the function dynamically constructs and solves a decision-tree version of the corresponding influence-diagram problem. Functions similar to MEV in Equation 4 have been developed previously, as for example to solve stochastic dynamic programming problems [Bellman57]. Thus, the general form of MEV in Equation 4 is not new. Rather, MEV is a specific version of the general form that provides a concrete bridge between belief network and influence diagram inference.

Although we have focused the discussion on the calculation of maximum expected value, other calculations can be performed using minor modifications of Equation 4, as for example sensitivity analyses and value lotteries. In addition, if an influence diagram is placed in Howard canonical form [Matheson88] then Equation 4 can be applied with only minor modification to calculate value of information.

3.3 Computational Efficiency

It is possible to significantly increase the efficiency of solving some types of influence-diagram problems. Here we consider two techniques that are used by Shachter's influence diagram algorithm [Shachter86a]. First, dynamic programming can be used, when the principle of optimality holds, to solve individual decision subproblems as a means of solving the global decision problem. In the best case, for binary decisions in $D_r$, this technique will decrease the computational time complexity by a factor on the order of $2^{|D_r|}$. Dynamic programming is equally applicable to efficiently solving belief network versions of influence diagram problems. Second, the time complexity may be reduced by effectively summing-out *only once* the intermediate uninstantiated nodes in an influence diagram. For example, in solving the decision problem for the influence diagram in Figure 2, node B can be effectively summed-out once to give a direct probabilistic relationship between A and C, thereby yielding a smaller influence diagram that can be used more efficiently to find the optimal decision. A similar summing-out process is applicable to solving the corresponding problem for the belief network in Figure 3. In general, it appears that the techniques for solving influence diagrams efficiently can be applied to solve efficiently belief-network versions of influence-diagram problems. Conversely, efficient algorithms exist for solving belief networks with particular topologies and, as discussed in Section 4, these algorithms can be used for solving influence diagrams that contain the same topological features.

## 4. Discussion

The close relationship between influence diagrams and belief networks, as discussed above, makes it apparent that previously developed belief network algorithms can be adapted readily to solve influence-diagram problems. For example, the message-passing algorithm of Pearl for belief-network inference can be used to solve efficiently many influence diagrams that are singly connected [Pearl86a]. Those belief-network algorithms that are designed for performing probabilistic inference using multiply-connected belief networks can be used to perform expected-value decision making with multiply-connected influence diagrams. One example of an applicable multiply-connected belief-network algorithm is the method of cutset conditioning, developed by Pearl [Pearl86b]. An adaptation of Pearl's algorithm [Suermondt88a] has been successfully applied to implement an influence diagram algorithm using the techniques in this paper [Suermondt88b]. Lauritzen and Spiegelhalter also have developed an algorithm for dealing with multiply-connected networks [Lauritzen88]. Their algorithm seems particularly efficient for performing inference using multiply-connected networks that have small clusters of nodes.

Unfortunately, inference using either belief networks or influence diagrams is NP-hard [Cooper87]. Therefore, for some complex, multiply-connected networks, it may be necessary to use *approximation algorithms*. Approximation algorithms produce an inexact, bounded solution, but guarantee that the exact solution is within those bounds. Several approximation algorithms have been developed recently to address the computational complexity of belief network inference. For example, algorithms have been developed that bound the goal probability that constitutes a belief-network inference problem [Cooper84, Peng87a, Henrion88]. These bounds are incrementally tightened as more computation time is expended. The application of these algorithms to influence-diagram problems leads to incremental tightening of the bounds on expected utility. For a decision to be made, sufficient computation must be expended so that the lower bound of some decision alternative is greater than the upper bounds of all other possible alternatives. The question then is whether such bounds can be made sufficiently tight in an acceptable amount of time for the inference problems in a particular domain.

Another type of belief network approximation method uses a Monte Carlo technique to produce a unique, point-valued probability estimate of some node of interest, plus a standard error of that estimate [Henrion86]. As more computation time is expended, the standard error of the probability decreases. When the algorithm is applied to an influence-diagram problem using Equation 4, it calculates an estimate of the expected-value of each decision alternative and a standard error of that estimate. For example, consider a modification of the influence diagram in Figure 2 where A is a disease, C is a binary-valued finding with known value T, $D_1$ is a treatment choice, and B is replaced by a complex pathophysiological network that represents the causal mechanisms by which disease A can cause the finding C. Suppose the immediate objective is to determine the treatment option that maximizes the expected value of the patient. Then, the Monte Carlo algorithm is applied in stages, with each stage corresponding to a different instantiation of treatment variable $D_1$. If the marginal probability of finding C is not too small, then for each treatment option the Monte Carlo method will rapidly converge on an estimate of the expected value which has a small standard error. The option with the maximum expected-value estimate is chosen as the optimal treatment. Note that the longer the Monte Carlo algorithm is applied, the smaller the standard error becomes. In contrast, if network B is highly connected secondary to complex pathophysiological interactions, exact influence diagram algorithms may require computation time that is exponential in the number of nodes in B in order to determine the expected value for each treatment option. The contrast in this case is between knowing the expected values exactly but after a long delay versus knowing only estimates of the expected values but knowing them quickly. Although Henrion's Monte Carlo method appears promising for some types of problems, it generally converges very slowly

61

when there are numerous evidence nodes or when the marginal probability of evidence is very small [Henrion86, Chin87]. Thus, the method is not always practical. Nonetheless, there may be numerous instances in which the application of approximate algorithms, such as Henrion's algorithm, is preferable to the application of exact algorithms.

Heuristic algorithms constitute another approach to the search for acceptable inference efficiency. *Heuristic belief-network algorithms*, as we use the term here, are not formally guaranteed to yield a correct probability in either the exact or the nontrivially bounded sense. Nonetheless, they may yield probabilities that are acceptably accurate. Furthermore, heuristic algorithms generally are fast, even in the worst case. For example, one potential approach involves converting belief networks to probabilistic neural networks [Geffner87, Peng87b] and using neural network inference algorithms to heuristically perform belief network inference.

Previous work in developing belief-network algorithms can be applied to solving influence-diagram problems. It seems important to analyze in detail, both theoretically and empirically, the relative efficiency of current belief-network algorithms and influence-diagram algorithms for solving various types of influence-diagram problems. This analysis may provide insights into designs for more efficient belief-network and influence-diagram algorithms. In addition, the design and analysis of such new algorithms may be facilitated by the relative uniformity of the belief-network representation and by the conceptual simplicity of the belief-network inference task.

ACKNOWLEDGEMENTS

I wish to thank Jaap Suermondt, Lyn Dupre, and the Workshop reviewers for valuable comments on this paper. This work has been supported in part by National Science Foundation grant IRI-8703710 and by National Library of Medicine grant LM-07033. Computer facilities were provided by the SUMEX-AIM resource under National Institutes of Health grant RR-00785.